\documentclass[letterpaper, 10 pt, conference]{ieeeconf}  

\IEEEoverridecommandlockouts                              

\title{\LARGE \bf
LCP-Fusion: A Neural Implicit SLAM with Enhanced Local Constraints and Computable Prior
}

\author{ Jiahui Wang, Yinan Deng, Yi Yang and Yufeng Yue$^{*}$
\thanks{This work is supported by the National Natural Science Foundation of China under Grant 92370203, 62233002. (*Corresponding Author: Yufeng Yue, yueyufeng@bit.edu.cn)}
\thanks{$^1$ School of Automation, Beijing Institute of Technology, Beijing, China.}
}

\usepackage{amsmath}
\usepackage{color}
\usepackage{amssymb}
\usepackage{graphicx}

\usepackage{float}
\usepackage{booktabs}
\usepackage{stfloats}
\usepackage{multirow}
\usepackage{cite}

\usepackage[colorlinks,
            linkcolor=red,
            anchorcolor=green,
            citecolor=green
            ]{hyperref}

\begin{document}

\maketitle
\thispagestyle{empty}
\pagestyle{empty}

\begin{abstract}

Recently the dense Simultaneous Localization and Mapping (SLAM) based on neural implicit representation has shown impressive progress in hole filling and high-fidelity mapping. Nevertheless, existing methods either heavily rely on known scene bounds or suffer inconsistent reconstruction due to drift in potential loop-closure regions, or both, which can be attributed to the inflexible representation and lack of local constraints. In this paper, we present LCP-Fusion, a neural implicit SLAM system with enhanced local constraints and computable prior, which takes the sparse voxel octree structure containing feature grids and SDF priors as hybrid scene representation, enabling the scalability and robustness during mapping and tracking. To enhance the local constraints, we propose a novel sliding window selection strategy based on visual overlap to address the loop-closure, and a practical warping loss to constrain relative poses. Moreover, we estimate SDF priors as coarse initialization for implicit features, which brings additional explicit constraints and robustness, especially when a light but efficient adaptive early ending is adopted. Experiments demonstrate that our method achieve better localization accuracy and reconstruction consistency than existing RGB-D implicit SLAM, especially in challenging real scenes (ScanNet) as well as self-captured scenes with unknown scene bounds. The code is available at \href{https://github.com/laliwang/LCP-Fusion}{https://github.com/laliwang/LCP-Fusion}.

\end{abstract}

\section{INTRODUCTION}

Dense visual Simultaneous Localization and Mapping (SLAM) plays a vital role during perception, navigation and manipulation in unknown environments. In recent decades, traditional SLAM methods \cite{deng2022s,mur2017orb, engel2017direct} have made significant progress in localization accuracy and real-time applications. However, due to the use of explicit scene representations like occupancy-grids \cite{hornung2013octomap, deng2023see, hd-ccsom}, point cloud \cite{deng2024opengraph, tang2023multi, tang2023ssgm}, Signed Distance Function \cite{oleynikova2017voxblox, newcombe2011kinectfusion, deng2024macim}, and surfels \cite{behley2018efficient}, which directly store and update limited scene information at fixed resolution without context, they struggle to balance memory consumption and mapping resolution, while being unable to reconstruct complete and consistent surfaces in noisy or unobserved areas.

Therefore, recent research has focused on implicit representation using neural networks \cite{mescheder2019occupancy} or radiance fields \cite{mildenhall2021nerf} to encode any points in scenes as continuous function, which can be used to extract isosurfaces at arbitrary resolution or to synthesize realistic unseen views. Utilizing the representation coherence and ability to render unseen views, numerous neural implicit SLAM systems \cite{sucar2021imap, zhu2022nice, wang2023co} have emerged to perform high-fidelity mapping and camera tracking in various scenes. However, most of them require known scene bounds due to inflexible scene representation \cite{yang2022vox}, resulting in performance degradation or failure in unknown scenarios.

\begin{figure}
    \centering
    \includegraphics[width=1\linewidth]{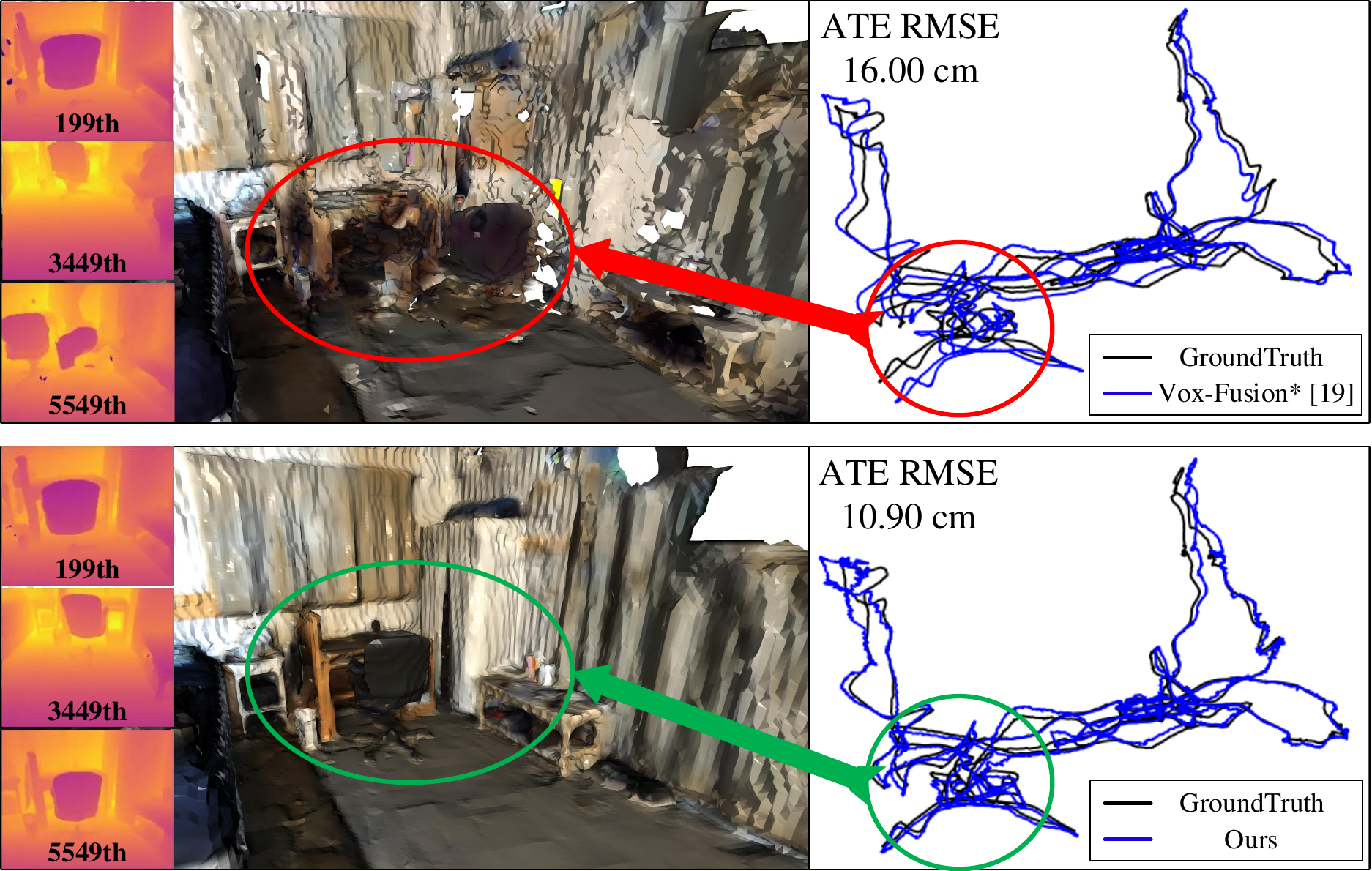}
    \caption{Example from baseline \cite{yang2022vox} of inconsistent surfaces due to drift in potential loop-closure regions composed of frames 119, 3449 and 5549 (top). Our method can reconstruct unknown scenes with less drift utilizing enhanced local constraints and easily computable SDF priors (bottom).}
    \label{Fig. 1}
\end{figure}

Focusing on applications in unknown scenes, one of the mainstream solutions is to allocate implicit feature grids dynamically in surface areas utilizing flexible sparse voxel octree (SVO) \cite{yang2022vox}. Since SVO-based methods \cite{vespa2018efficient} only represent scenes using high-dimensional features in sparse voxel grids, they tend to be sensitive to odometry drift caused by their insufficient local constraints in loop closure regions. This may contribute to inconsistent reconstruction much further, which is shown in Fig. \ref{Fig. 1}. Additionally, with an explicit SDF octree prior, hybrid methods \cite{jiang2023h} are proposed for precise mapping, but instead use a traditional visual odometry \cite{qin2018vins} as tracking module. Thus, for the unified dense SLAM system utilizing a neural implicit representation for both tracking and mapping, it is worth investigating in alleviating reconstruction inconsistency caused by localization drift in unknown scenes with potential loop closure.

To this end, we introduce LCP-Fusion (a neural implicit SLAM system with enhanced \textbf{L}ocal \textbf{C}onstraints and \textbf{C}omputable \textbf{P}rior), which can alleviate drift in potential loop-closure without other external modules. Our key ideas are as follows. \textbf{First}, to handle unknown scene boundaries, we utilize the SVO to dynamically allocate hybrid voxel grids containing coarse SDF priors and residual implicit features, which yields scene geometry and color through sparse volume rendering. \textbf{Second}, through pixel reprojection between frames, we propose a novel sliding window selection strategy based on visual overlap, which not only strengthens local constraints but also alleviates catastrophic forgetting. In addition to only evaluating individual frame, a practical warping loss constraining relative poses is introduced to further improve localization accuracy. \textbf{Third}, to reduce the redundancy of iterations in joint optimization, we adopt an adaptive early ending without significant performance degradation owing to our proposed hybrid representation. We perform extensive evaluations on a series of RGB-D sequences to demonstrate the localization improvement of our method, as well as the applications in real scenes with unknown bounds. In summary, our contributions are:

\begin{itemize}
    \item We present LCP-Fusion, a neural implicit SLAM system based on hybrid scene representation, which allocates hybrid voxels with implicit features and estimated SDF priors dynamically in scenes without known bounds.
\end{itemize}

\begin{itemize}
    \item  We introduce a novel sliding window selection strategy based on visual overlap and a warping loss constraining relative poses for the enhanced local constraints.
\end{itemize}

\begin{itemize}
    \item  Extensive evaluations on various datasets demonstrate our competitive performance in localization accuracy and reconstruction consistency, as well as robustness to fewer iterations and independence on scene boundaries.
\end{itemize}
 
\section{RELATED WORKS}

While existing neural implicit SLAM systems demonstrate impressive performance in high-fidelity reconstruction, incremental consistent mapping in unknown scenes remains challenging. We mainly attribute this to the following aspects: 1) inflexible implicit scene representation; and 2) insufficient local constraints during optimization.

\subsection{Implicit Scene Representation}

Compared to earlier works \cite{mescheder2019occupancy, park2019deepsdf} requiring 3D ground truth for supervision in object-level reconstruction, NeRF and its variants \cite{mildenhall2021nerf, yue2024lgsdf} encode 3D points in the weights of MLP directly from image sequences. Against earlier works \cite{lionar2021neuralblox,peng2020convolutional} that required well-designed encoder-decoder networks to capture scenes, NeRF-based SLAM jointly optimizes implicit maps and camera poses from end to end using shallow MLP decoders. While iMAP \cite{sucar2021imap} first introduces NeRF \cite{mildenhall2021nerf} into the framework of dense SLAM, due to the limited capability and slow training speed of a single MLP, its mapping and tracking effects drop during continual learning. Thus, Nice-SLAM \cite{zhu2022nice} proposes to encode local scenes with multi-resolution dense feature grids and pre-trained MLP decoder \cite{peng2020convolutional}, which greatly enriches mapping details and speeds up the training with less catastrophic forgetting. Furthermore, Co-SLAM \cite{wang2023co} designs a joint encoding to combine the benefits of coordinate encoding and parametric embedding, achieving both consistent and sharp real-time reconstruction.

However, these methods above assume known scene bounds in application, as they have a requirement for scene bounds to allocate dense grids or perform normalized positional encoding before running. To this end, Vox-Fusion \cite{yang2022vox} utilizes sparse voxel octree (SVO) to allocate feature grids on-the-fly and force scene learning towards surface areas. Since \cite{yang2022vox} only encodes scene at the scale of leaf voxel, the feature drift caused by odometry drift is difficult to correct in high dimensional space. Thus, we adopt a hybrid scene representation that store an easy-to-get SDF prior at the same resolution, which can not only provide reasonable initialization but also facilitate the correction with explicit updates in low dimensional space.

\subsection{Local Constraints during Optimization}

To address reconstruction consistency during incremental mapping, most implicit SLAM systems jointly optimize implicit maps and keyframe poses \cite{tang2023robust} with a sliding window, while some introduce additional relative pose constraints. 

For sliding window selection, iMAP \cite{sucar2021imap} and iSDF \cite{ortiz2022isdf} select keyframes according to their loss distribution, which is time consuming during active sampling. With pre-trained MLP decoder, Nice-SLAM \cite{zhu2022nice} only selects keyframes overlapped with the current frame without considering catastrophic forgetting. For Vox-Fusion \cite{yang2022vox}, without pre-trained priors, it has to randomly select keyframes from the global keyframe set, potentially weakening local constraints between overlapped frames. As for Co-SLAM \cite{wang2023co}, they randomly sample rays from the global keyframe set for global BA, which is optimized across all keyframes yet lack of local constraints. In summary, most sliding windows lack sufficient local constraints from overlapped loop-closure frames.

For relative pose constraints, Nicer-SLAM \cite{zhu2024nicer} introduces a sophisticated RGB warping loss among random and latest keyframes to further enforce geometry consistency, and Nope-Nerf \cite{bian2023nope} performs warping process between every consecutive frames to constrain relative poses. However, they only constrain relative poses between nearby or random frames instead of frames with visual overlap in loop closure.

Concurrently, while DIM-SLAM \cite{li2023dense} (partially open source) employs a similar strategy to enhance local constraints, it relies exclusively on RGB inputs and multiple levels of dense feature grids. This approach may result in less precise reconstruction and increased reliance on environmental priors such as scales. Additionally, although other recent methods \cite{johari2023eslam,zhang2023go} are meticulously crafted for loop closure, they are still constrained by known scene boundaries. 

To this end, we further introduce warping loss for relative pose constraints between potential loop-closure frames, in a sliding window based on visual overlap.

\begin{figure*}
    \centering
    \includegraphics[width=\linewidth]{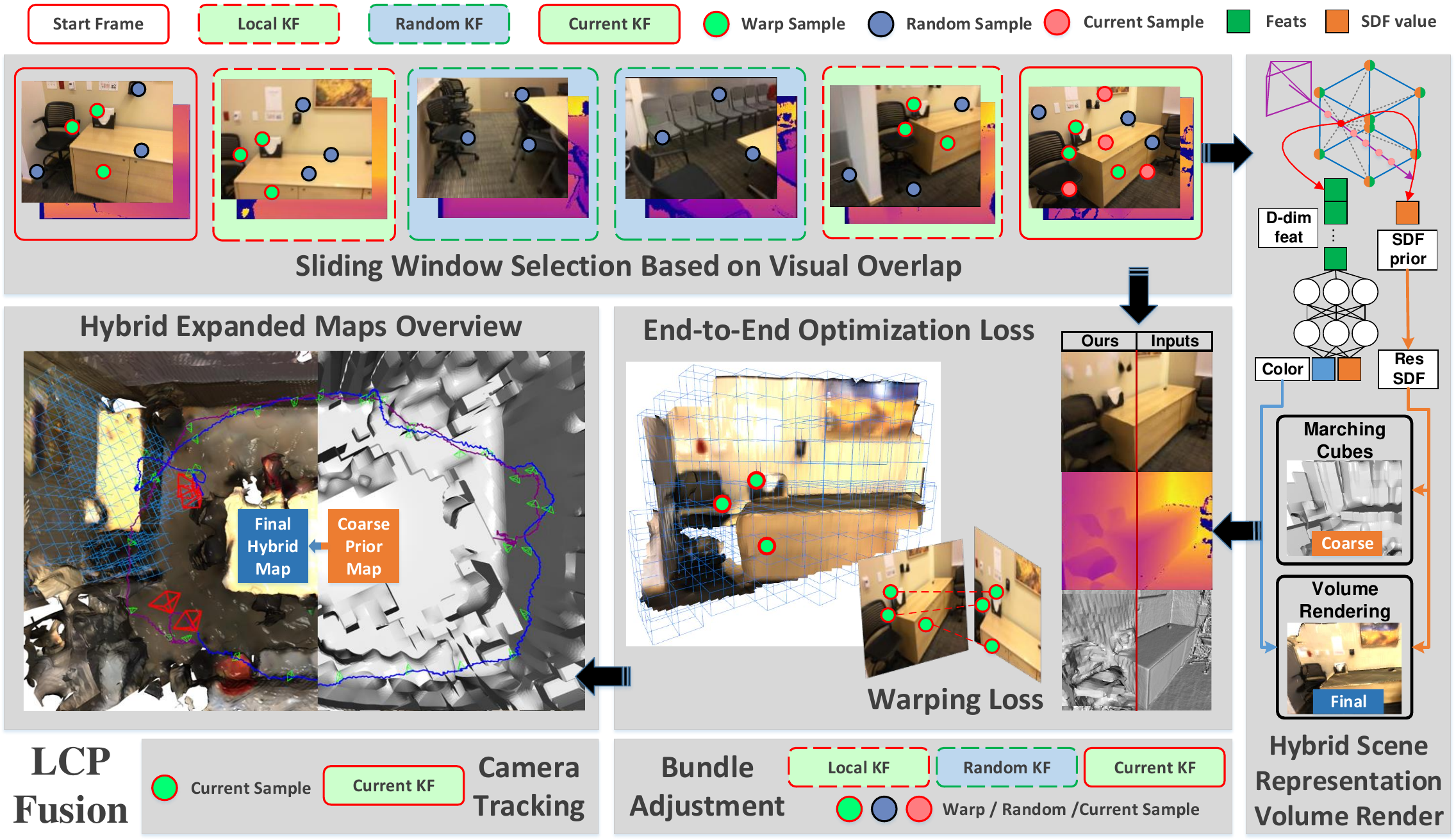}
    \caption{Overview of our LCP-Fusion system. Receiving RGB-D inputs with initialized poses from the tracking process, we jointly optimize hybrid scene representation and camera poses among our sliding window, which contains more visual overlap with the current frame in potential loop closure regions. Additionally, the proposed warping loss can obtain sufficient constraints for relative poses with visual overlap.}
    \label{overview}
    \vspace{-1.0em}
\end{figure*}

\section{LCP-Fusion} \label{LCP-Fusion}

An overview of our system is shown in Fig. \ref{overview}. Receiving continuous RGB-D frames of color images \(\boldsymbol{C}_{i}^{obs}\in \boldsymbol{R}^{3}\) and depth images \(\boldsymbol{D}_{i}^{obs}\in \boldsymbol{R}\) without poses, our expanded hybrid representation first allocates hybrid voxels (Sec. \ref{rep}) where exists valid point clouds under estimated pose from the tracking process. 
Then volume rendering based on ray-voxel intersection is processed through SDF prior and implicit feature grids, which yields rendered RGB \(\hat{\boldsymbol{C}_{i}}\) , Depth \(\hat{\boldsymbol{D}_{i}}\) and predicted SDF values \(s_{i}\) for optimization. By employing a sliding window selection strategy based on visual overlap (Sec. \ref{window}), more keyframes related to the current frame are evaluated during bundle adjustment optimization. Then several loss functions include our warping loss are defined to optimize camera poses and scene representation during tracking and mapping (Sec. \ref{optim}).

\subsection{Hybrid Scene Representation } \label{rep}
In general, we represent unknown scene utilizing a hybrid multi-level SVO to allocate leaf voxels \(v^i\) dynamically,  which store D-dim feature embeddings \({\left \{  \boldsymbol{e}_{k=1\sim 8} \right \}}_i \) and 1-dim SDF priors \({\left \{  s_{k=1\sim 8}^{prior}\right \}}_i \) at each vertex, where SDF priors are easily computed through back-projection and provided as reasonable initialization for implicit features.

During volume rendering, sampled point \(\boldsymbol{p}_j\) along rays can easily get point-wise features \(\boldsymbol{E}_j\) and coarse SDF values \(s_{j}^{c} \) through trilinear interpolation \(TriLerp\left (  \cdot \right )\). Then \(\boldsymbol{E}_j\) can be sent to MLP decoder \(\boldsymbol{M}_{\theta } \left ( \cdot  \right )\) for residual SDF \(s_{j}^{res}\in R\) and color \(\boldsymbol{c}_{j}\in \boldsymbol{R}^{3}\) at point \(\boldsymbol{p}_j\). Along with \(s_{j}^{c} \) from SDF priors, final results for volume rendering can be obtained:
\begin{equation}
(\boldsymbol{E}_{j},s_{j}^{c})=TriLerp\left ( \boldsymbol{p}_{j},\left \{\boldsymbol{e}_{k=1\sim 8},s_{k=1\sim 8}^{prior}   \right \}_i   \right ) \\
\end{equation}
\begin{equation}
    \left ( \boldsymbol{c}_{j},s_{j}^{res}   \right )=\boldsymbol{F}_{\theta }\left (  \boldsymbol{E}_{j} \right ),s_{j}=s_{j}^{c}+s_{j}^{res}
\end{equation}
\par
\begin{figure}
    \centering
    \includegraphics[width=1\linewidth]{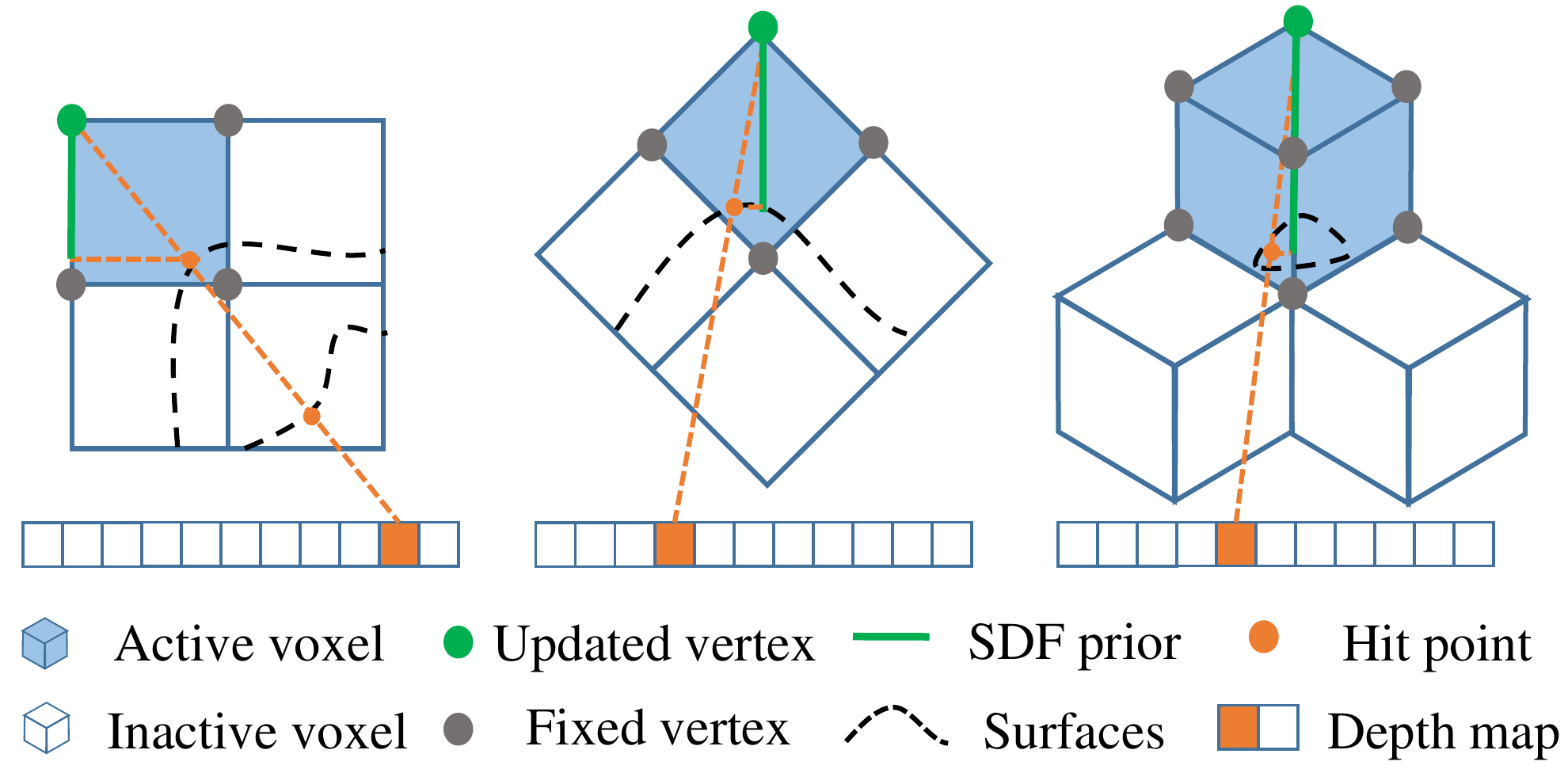}
    \caption{Visualization of SDF prior estimates. To avoid unreasonable SDF priors due to occlusion, we indicate three extreme cases  from left to right.}
    \label{fig:SDF}
     \vspace{-1.5em}
\end{figure}

 As shown in Fig. \ref{fig:SDF}, whenever enough points from new frames drop inside the leaf voxel \(v^i\), we first estimate current SDF priors \(s_{curr}^{prior}\) at each vertex through projecting them onto current depth frame, and then update them separately like TSDF fusion \cite{newcombe2011kinectfusion}:
\begin{equation}
    s_{curr}^{prior}=\boldsymbol{D}(\boldsymbol{u})-d_{\boldsymbol{p}}
\end{equation}
\begin{equation}
    \left\{\begin{matrix}
 s_{fusion}^{prior}=\frac{s_{fusion}^{prior} \cdot n_{update} + s_{curr}^{prior} \cdot n_{curr}}{n_{update}+n_{curr}}  \\
n_{update}=n_{update}+n_{curr}
\end{matrix}\right.
\end{equation}
where \(d_{\boldsymbol{p}}\) is the z-axis distance of vertex \(\boldsymbol{p}\) under the current camera coordinate, and \(\boldsymbol{D}(\boldsymbol{u})\) is the depth value at the re-projected pixel \(u\). Due to the fact that a vertex can be shared by up to 8 surrounding voxels, \(s_{curr}^{prior}\) tends to be updated by several voxels. We directly take updated voxels number \(n_{curr}\) as current update weight to obtain final values \(s_{fusion}^{prior}\) through weighted fusion while updating \(n_{update}\).

However, before updating \(s_{fusion}^{prior}\) with current estimate \(s_{curr}^{prior}\), we need to eliminate the invalid estimates in the following situations:
\begin{enumerate}
    \item The re-projected pixel \(\boldsymbol{u}\) of vertex \(\boldsymbol{p}\) locates outside current frame.
    \item The depth value \(\boldsymbol{D}(\boldsymbol{u})\) of pixel \(\boldsymbol{u}\) is absent due to sensor noise or depth truncation.
    \item The estimate value \(s_{curr}^{prior}\) cannot meet \( \left | \boldsymbol{D}(\boldsymbol{u})-d_{\boldsymbol{p}} \right | < \sqrt{3} \times (voxel\ size)\) as Fig. \ref{fig:SDF}.
\end{enumerate}
\par With explicit initialization and updates in low dimensional space, although at the same resolution, our system achieves improved accuracy in localization and stability over fewer iterations of optimization, as well as consistent reconstruction in noisy or inaccurate areas.

Much further,  receiving inferred values \((\boldsymbol{c}_j,s_j)\) of \(N\) points from hybrid representation, sparse volume rendering performs along each ray as follows:
\begin{equation}
\begin{aligned}
w_j&=\sigma (\frac{s_j}{tr})\cdot \sigma (-\frac{s_j}{tr})\\
\hat{\boldsymbol{C}}=\frac{1}{ {\textstyle \sum_{j=0}^{N-1}}w_j }\sum_{j=0}^{N-1}& w_j \cdot \boldsymbol{c}_j,\ \hat{D}=\frac{1}{ {\textstyle \sum_{j=0}^{N-1}}w_j }\sum_{j=0}^{N-1} w_j \cdot d_j    
\end{aligned}
\end{equation}
where \(\sigma(\cdot)\) is the sigmoid function and \(tr\) is a pre-defined truncation distance. With the point-wise rendering weight \(w_j\) computed by predicted SDF \(s_j\), rendered RGB \(\hat{\boldsymbol{C}}\) and depth \(\hat{D}\) of rays are obtained by weighting predicted point-wise color \(\boldsymbol{c}_j\) and sampled depth \(d_j\) respectively. The rendered RGB images, depth images and predicted SDF values will be used to evaluate with inputs for optimization in Sec. \ref{optim}.
\subsection{Sliding Window Selection} \label{window}

\begin{figure}
    \centering
    \includegraphics[width=\linewidth]{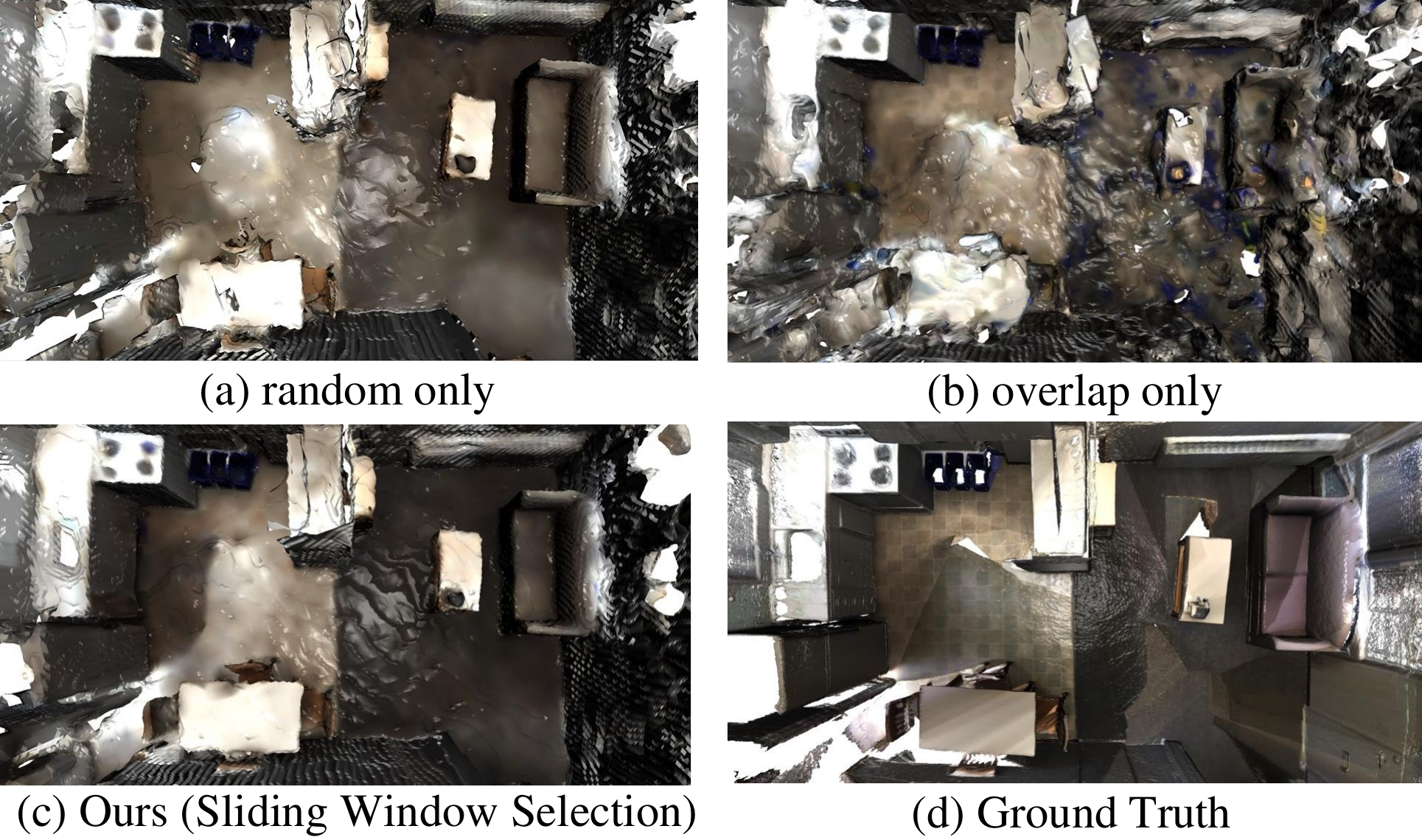}
    \caption{Aerial view of reconstruction on scene0181 \cite{dai2017scannet} using different sliding window designs. Through comparison, our sliding window selection method can not only improve localization but also alleviate forgetting. }
    \label{fig:overlap_comp}
     \vspace{-1.5em}
\end{figure}

For selection of optimized keyframe in sliding windows, we propose a light yet efficient strategy to enhance local constraints and alleviate catastrophic forgetting as well. Specifically, our sliding window \(F_{window}\) consists of three parts: local keyframes \(F_{local}\), historical keyframes \(F_{his}\) and current frame \(F_{curr}\). For current frame \(F_{curr}\) with optimized camera pose, we first randomly sample \(N_{rep}\) pixels \(\boldsymbol{q}_c\) with valid depth values, and then re-project them onto every keyframe in global keyframe list \({\textstyle \sum_{i}^{N_{kf}}}F_{{kf}_i}\) for re-projected pixels \(\boldsymbol{q}_{{kf}_i}\). After counting the number of re-projected pixels that have successfully dropped inside each keyframe as \({Count}_{{kf}_i}\), we can obtain each component of sliding windows as follows:

\begin{equation}
\begin{aligned}
F_{local}=\mathop{maxsort}\limits_{{Count}_{{kf}_i}}( {\textstyle \sum_{i}^{N_{kf}}}F_{{kf}_i}, \mathcal{W}/2) \\
F_{his}=\mathop{randsort}( {\textstyle \sum_{i}^{N_{kf}}}F_{{kf}_i}, \mathcal{W}/2) \\
F_{window}=F_{local}+F_{his}+F_{curr}
\end{aligned}
\end{equation}
where \(\mathcal{W}+1\) is the width of sliding window, local keyframes \(F_{local}\) are selected among keyframes that have most points projected onto, historical keyframes \(F_{his}\) are selected randomly from global keyframe list. Both of them are essential for enhancing local constraints and alleviating forgetting, which is shown in Fig. \ref{fig:overlap_comp}. Especially, in scenes with frequent loop closures, \(F_{local}\) are modified to be selected randomly among \(2\mathcal{W}\) keyframes with most re-projected points, obtaining adequate constraints at far enough intervals.
\subsection{End to End Optimization} \label{optim}
1) \textit{Loss Functions: }To supervise the learning of camera poses and hybrid representation, we apply the similar loss functions as \cite{yang2022vox}: re-render RGB loss (\({\mathcal{L}}_{rgb}\)) and Depth loss (\({\mathcal{L}}_{d}\)), SDF free-space loss (\({\mathcal{L}}_{fs}\)) and estimate loss (\({\mathcal{L}}_{sdf}\)) on a batch of rays \((R=R_{t,m})\) sampled randomly from current frame or sliding windows:
\begin{equation}
\begin{aligned}
    \mathcal{L}_{rgb}=\frac{1}{\left | R \right | } &\sum_{r\in R} \left \| \hat{\boldsymbol{C}_{r}}-\boldsymbol{C}_{r}^{obs} \right \|, \ \mathcal{L}_{d}=\frac{1}{\left | R \right | } \sum_{r\in R} \left \| \hat{D_{r}}-D_{r}^{obs} \right \| \\
&\mathcal{L}_{fs}=\frac{1}{\left | R \right | } \sum_{r\in R} \frac{1}{N_{r}^{fs}} \sum_{n\in N_{r}^{fs}}(s_{n}-tr)^2 \\
\mathcal{L}_{sdf}&=\frac{1}{\left | R \right | } \sum_{r\in R} \frac{1}{N_{r}^{tr}} \sum_{n\in N_{r}^{tr}}(s_{n}-(D_{r}^{obs}-d_n))^2
\end{aligned}
\end{equation}
where photometric loss \({\mathcal{L}}_{rgb}\) and \({\mathcal{L}}_{d}\) are defined between render results \((\hat{\boldsymbol{C}_{r}},\hat{D_{r}})\) and observations \((\boldsymbol{C}_{r}^{obs},D_{r}^{obs})\); geometirc loss \({\mathcal{L}}_{fs}\) and \({\mathcal{L}}_{sdf}\) are defined respectively on point-wise predicted SDF values \(s_{n}\) outside and inside truncation distance along rays. 

Additionally, during bundle adjustment optimization, we introduce a warping loss defined between current frame and sliding window keyframes on another batch of rays \(R_w\) to constrain relative poses and enforce geometry consistency. Let \(\boldsymbol{q}_c\) denotes a 2D pixel with valid depth value at current frame \(F_{curr}\), we first project it into 3D space with camera parameters \((\boldsymbol{K},{\tilde{\boldsymbol{R}}}_c|{\tilde{\boldsymbol{t}}}_c) \) and then re-project it onto sliding window keyframes \(F_{w\in{\mathcal{W}}}\) as:
\begin{equation}
\boldsymbol{q}_w=\boldsymbol{K} \widetilde{\boldsymbol{R}}_w{ }^T\left(\widetilde{\boldsymbol{R}}_c \boldsymbol{K}{ }^{-1} \boldsymbol{q}_c^{\text {homo }} D_{\boldsymbol{q}_c}+\tilde{\boldsymbol{t}}_c-\tilde{\boldsymbol{t}}_w\right)
\end{equation}
where \(\boldsymbol{q}_{c}^{homo}={(u,v,1)}^T\) is the homogeneous coordinate of \(\boldsymbol{q}_c\) and \(D_{\boldsymbol{q}_c}\) denotes its depth observation. With the aid of depth observation and visual overlap in sliding windows, we define the RGB and depth warping loss respectively on the current frame:
\begin{equation}
    \begin{aligned}
        &\mathcal{L}_{{warp}_{R,D}}=\frac{1}{|R_w|}\sum_{r\in R_w}^{} \sum_{w\in \mathcal{W},w\ne c}^{} \left | {(\boldsymbol{I},D)}_{\boldsymbol{q}_c}^{r}-{(\boldsymbol{I},D)}_{\boldsymbol{q}_w}^{r} \right |
    \end{aligned}
\end{equation}
\par It is worth noting that since the sliding window defined in Sec. \ref{window} has more visual overlap with the current frame, we can obtain more valid re-projected points for warping constraints. The final loss function is a weighted sum of all above losses:
\begin{equation}
\begin{aligned}
    \mathcal{L}(\boldsymbol{P})={\lambda}_{rgb}{\mathcal{L}}_{rgb}+{\lambda}_{d}{\mathcal{L}}_{d}+{\lambda}_{sdf}{\mathcal{L}}_{sdf}\\
    +{\lambda}_{fs}{\mathcal{L}}_{fs}+{\lambda}_{{warp}_R}{\mathcal{L}}_{{warp}_R}+{\lambda}_{{warp}_D}{\mathcal{L}}_{{warp}_D}
\end{aligned}
\end{equation}
where \(\boldsymbol{P}=\left \{ \mathcal{\theta},\boldsymbol{E},\left \{ {\boldsymbol{\xi} }_t \right \} \right \}\) represent trainable MLP weights \(\theta\), feature grids \(\boldsymbol{E}\) and camera poses \(\left \{ exp({{\boldsymbol{\xi}}_t}^{\wedge }) \in SE(3) \right \}\). And we consider the above weights as constants without special declaration.

\par 2) \textit{Tracking and Mapping}: For Tracking, \(\boldsymbol{P}= \left \{ {\boldsymbol{\xi}}_{curr} \right \}\) and we employ final loss function without warping loss. For every new coming RGB-D frame \(F_{curr}\), we initialize the pose of \(F_{curr}\) identical to last tracked frame, and then sample \(R_t\) pixels randomly for volume rendering in frozen representation. Finally we only optimize current pose \({\boldsymbol{\xi}}_{curr} \in \mathfrak{s e}(3)\) through loss back-propagation for \({Iter}_t\) iterations.

For Mapping, \(\boldsymbol{P}=\left \{ \mathcal{\theta},\boldsymbol{E},\left \{ {\boldsymbol{\xi} }_t \right \} \right \}\) and we employ the entire final loss function. Receiving every tracked frame with coarse pose estimates, we sample \(R_m\) pixels among sliding window \(F_{window}\) and additional \(R_{w}\) pixels on current frame \(F_{curr}\) for joint optimization and warping constraints respectively. Finally we jointly optimize scene representation and camera poses of sliding window for \({Iter}_m\) iterations.

Moreover, we provide an adaptive early ending policy for mapping process, which encourages system to optimize more iterations at areas with higher loss while less iterations for others. The details and effects are demonstrated in Sec. \ref{exp_ablation}
\section{EXPERIMENT} \label{exp}

To evaluate the performance of our proposed method, we first compare its localization accuracy and reconstruction consistency with other RGB-D Neural implicit SLAM on the real-world ScanNet dataset \cite{dai2017scannet} , synthetic Replica dataset \cite{straub2019replica} and our self-captured dataset with a depth camera mounted on mobile robots. And then we demonstrate our independence of scene bounds against SOTA bounded methods \cite{wang2023co}, as well as the effects of our hybrid representation against \cite{yang2022vox}. Moreover, we also conduct ablation studies to confirm the effectiveness of each module from our method.

\subsection{Experimental Setup} \label{exp_set}
1)\textit{ Implementation Details}: For scene representation, we maintain a 8-level SVO to allocate leaf voxels of 0.2m, which store 16-D feature and 1-D SDF prior at each vertex. For MLP decoder, we use the same structure as \cite{yang2022vox} for residual SDF and color. \((R_{rep},R_w,R_t,R_m)=1024\) pixels are sampled randomly from each frame for optimization during tracking and mapping, \({Iter}_t=30\) and \({Iter}_m=15\). For mapping, we maintain a sliding window of \(\mathcal{W}=4\) and insert a new keyframe at fixed interval of 50. All experiments are carried out on a desktop system with an Intel Xeon Platinum 8255C 2.50GHz CPU and an NVIDIA Geforce RTX 3090 GPU.

2) \textit{Baselines}: We consider Nice-SLAM \cite{zhu2022nice}, Co-SLAM \cite{wang2023co} and Vox-Fusion \cite{yang2022vox} as our baselines for localization accuracy. Since \cite{zhu2022nice} and \cite{wang2023co} both require known scene bounds in advance, we provide 3D bounds for them following their instruction. Moreover, in order to demonstrate their dependence on known scene bounds, we provide coarse scene bounds for \cite{wang2023co} in Sec. \ref{scene bounds}, which can be oversized or undersized. And we reproduce \cite{yang2022vox} as Vox-Fusion* from their official release for fair comparison.

3) \textit{Metrics}: For evaluation of localization accuracy, we adopt absolute trajectory error (ATE) between estimate poses \(\left \{ P_{1},P_2,...,P_n \in SE(3)  \right \}\) and ground truth poses \(\left \{ Q_{1},Q_2,...,Q_n \in SE(3)  \right \}\). Specifically, we take the ATE RMSE for evaluation using scripts provided by \cite{yang2022vox}.

\subsection{Evaluation of Localization} \label{exp_result}

\begin{figure}
    \centering
    \includegraphics[width=\linewidth]{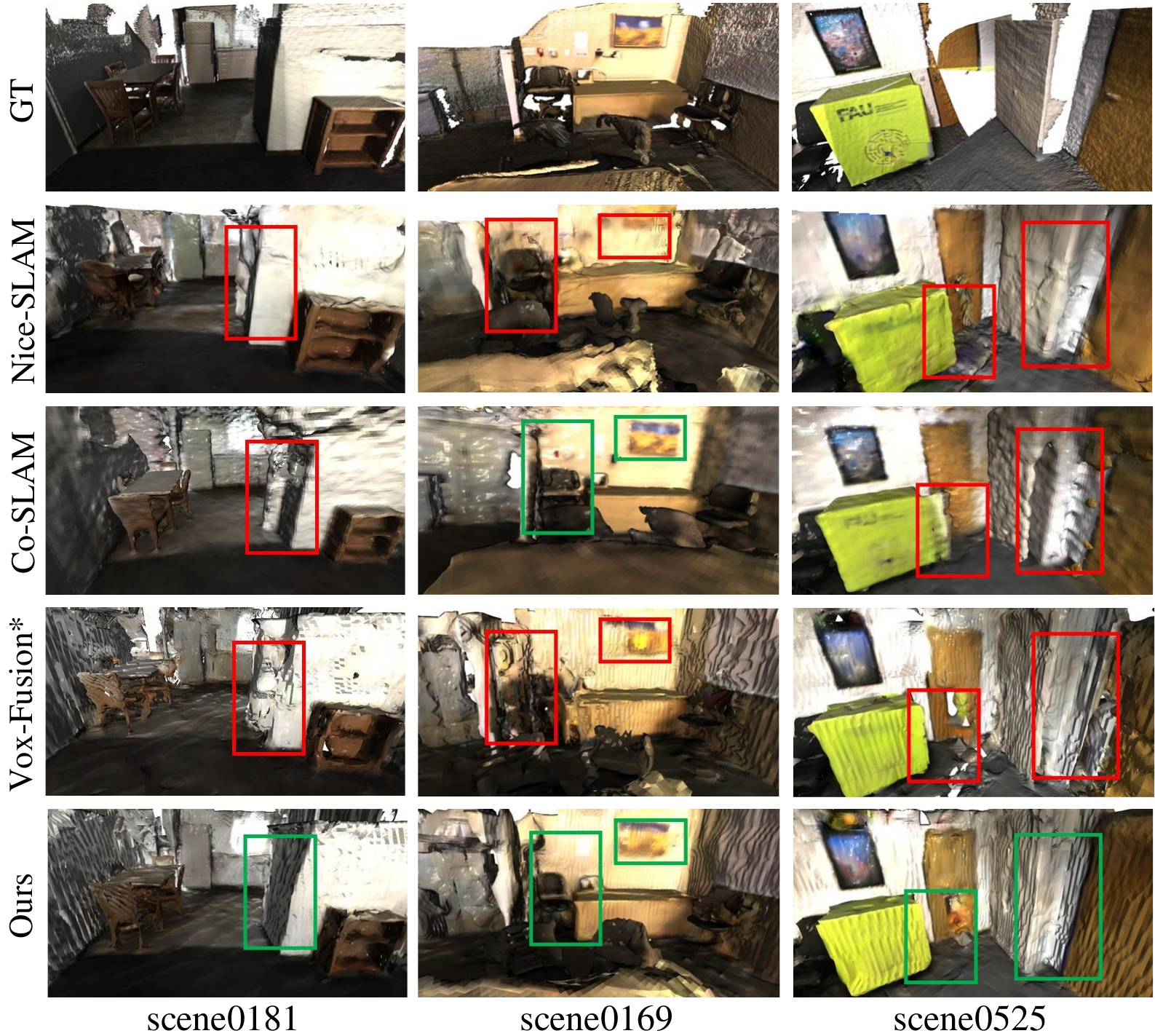}
    \caption{Reconstructed mesh results of potential loop-closure regions in ScanNet. For comparison, we highlight the regions with inconsistent surfaces by red boxes, while green boxes for consistent surface.}
    \label{mesh_comp}
    \vspace{-1.5em}
\end{figure}

\begin{figure}
    \centering
    \includegraphics[width=\linewidth]{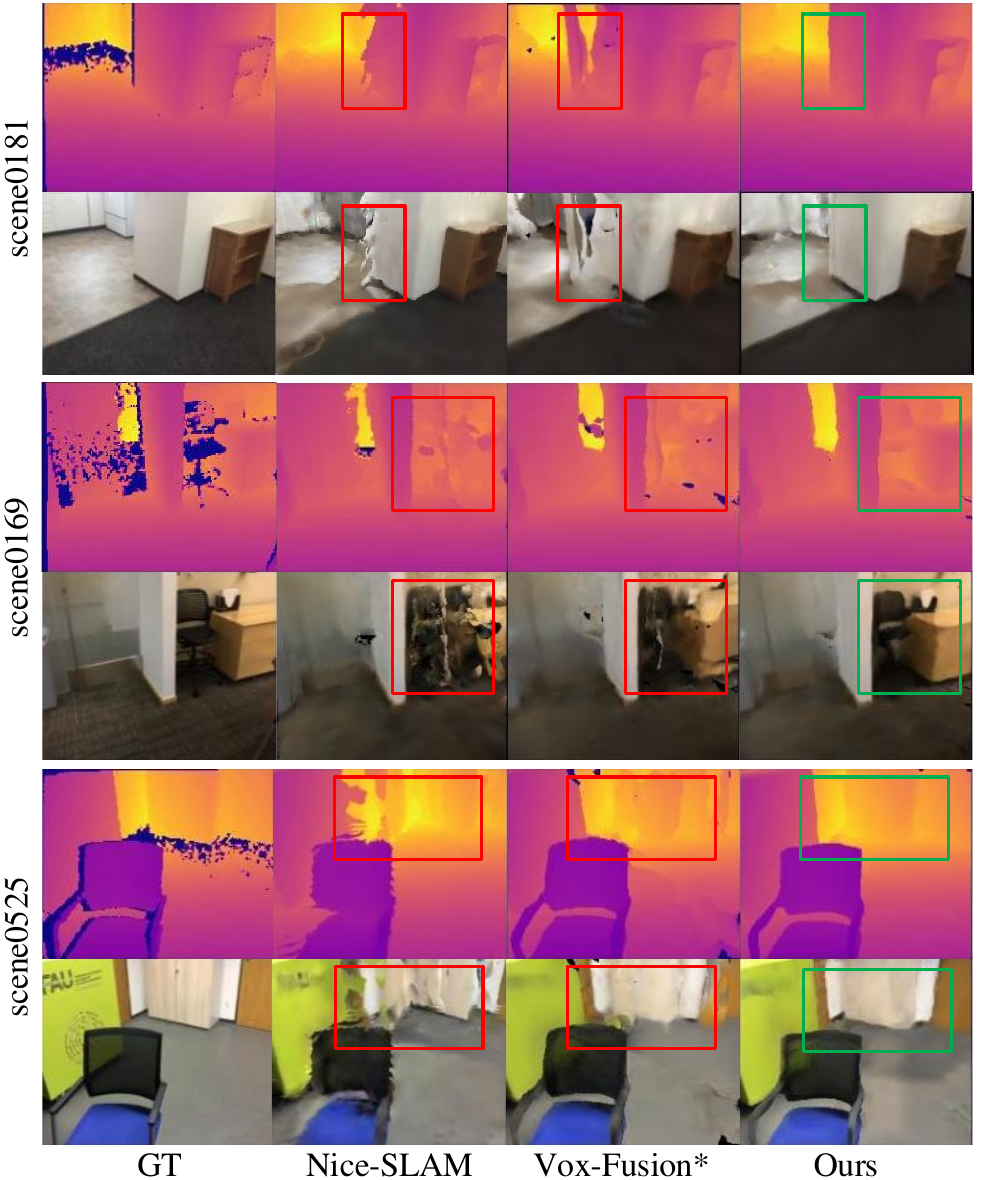}
    \caption{Rendered Depth and Color results of potential loop-closure regions in ScanNet. Our method can improve surface consistency in highlight boxes due to enhanced local constraints and SDF priors.}
    \label{render_comp}
    \vspace{-1.5em}
\end{figure}

1) \textit{Evaluation on ScanNet}: To demonstrate the improvement in localization accuracy, especially in challenging scenarios from real capture. We first evaluate our system on representative sequences of ScanNet, which are captured from real scenes with potential loop-closure and contain more noises in depth inputs. The quantitative results are shown in Table. \ref{table 1}. As can be seen, our localization accuracy surpass both bounded and unbounded methods by a large margin in these challenging scenes with loop closure. Moreover, we also qualitatively compare our system on reconstruction and render quality with baselines, shown in Fig. \ref{mesh_comp} and Fig. \ref{render_comp}. It can be seen that owing to our localization improvement brought by local constraints and priors, there are less inconsistent surfaces in scenes, which is common in previous reconstructions, such as the wall in scene0181, the desk in scene0169 and the wardrobe in scene0525.

\begin{table}[t]
\setlength{\tabcolsep}{2.5pt}
\caption{ATE RMSE (cm) on selected ScanNet sequences}
\label{table 1}
  \centering
    \begin{tabular}{ccccccccc}
   \toprule
   Scene ID& Reference& 0059 & 0169 & 0181 & 0207 & 0472 & 0525& Avg.\\
   \midrule
   Nice-SLAM & CVPR2022& 12.25 & 10.28 & 12.93 & 6.65 & 9.64& 10.31 & 10.34 \\
   Co-SLAM & CVPR2023& 12.29 & 6.62 & 13.43 &  7.13 &  12.38& 11.74& 10.60 \\
   Vox-Fusion* & ISMAR2022&9.06 & 9.64 & 15.38 & 7.74 &  9.98& 8.42& 10.04 \\
   Ours&  -& \textbf{7.56}& \textbf{5.91}& \textbf{10.18}&  \textbf{6.29}& \textbf{8.33}& \textbf{5.06}& \textbf{7.22}\\
   \bottomrule
\end{tabular}
\vspace{-1.0em}
\end{table}

2) \textit{Evaluation on Replica}: We also evaluate our localization enhancement on synthetic RGB-D sequences of Replica. Since the image quality of this synthetic dataset is particularly pure, and each image can provide sufficient geometric and color constraints for tracking and mapping, our baselines have achieved high localization accuracy. Even so, our system can still outperform them to some extent, due to our additional constraints during bundle adjustment optimization. The quantitative results are shown in Table. \ref{tabel 2}.

\begin{table}[t]
\setlength{\tabcolsep}{1.3pt}
\caption{ATE RMSE (cm) on selected Replica sequences}
\label{tabel 2}
  \centering
    \begin{tabular}{ccccccccc}
   \toprule
   Scene ID& Room0& Room1 & Room2 & Office1 & Office2 & Office3 & Office4& Avg.\\
   \midrule
   Nice-SLAM & 1.69& 2.04& 1.55 & 0.90 &1.39 &3.97&3.08 & 2.09 \\
   Co-SLAM & 0.67& 1.44 & 1.11 & \textbf{0.57 }& 2.10 & 1.58& 0.90& 1.20 \\
   Vox-Fusion* & 0.64&1.36 & 0.84 & 1.15 & 0.98& 0.72& 0.92& 0.94 \\
   Ours&  \textbf{0.54}& \textbf{1.02}& \textbf{0.78}& 1.08& \textbf{0.92}& \textbf{0.66}&\textbf{0.85}& \textbf{0.84}\\
   \bottomrule
\end{tabular}
\vspace{-1.0em}
\end{table}

3) \textit{Evaluation on Self-Captured}: To evaluate our performance in practical scenes with unknown bounds, we capture two RGB-D sequences by Azure Kinect camera mounted on mobile robots, which contains more missing depth values than ScanNet \cite{dai2017scannet}. Without known scene bounds in advance, we only compare our LCP-Fusion against Vox-Fusion* \cite{yang2022vox} for they both utilize extensible SVO-based scene representations. The quantitative and qualitative results shown in Fig. \ref{self_captured} demonstrate that our method achieve better localization accuracy and reconstruction consistency in inaccurate or noisy regions, such as the incomplete wall in sc601 and calibration board in sc614.

\begin{figure}[t]
    \centering
    \includegraphics[width=0.95\linewidth]{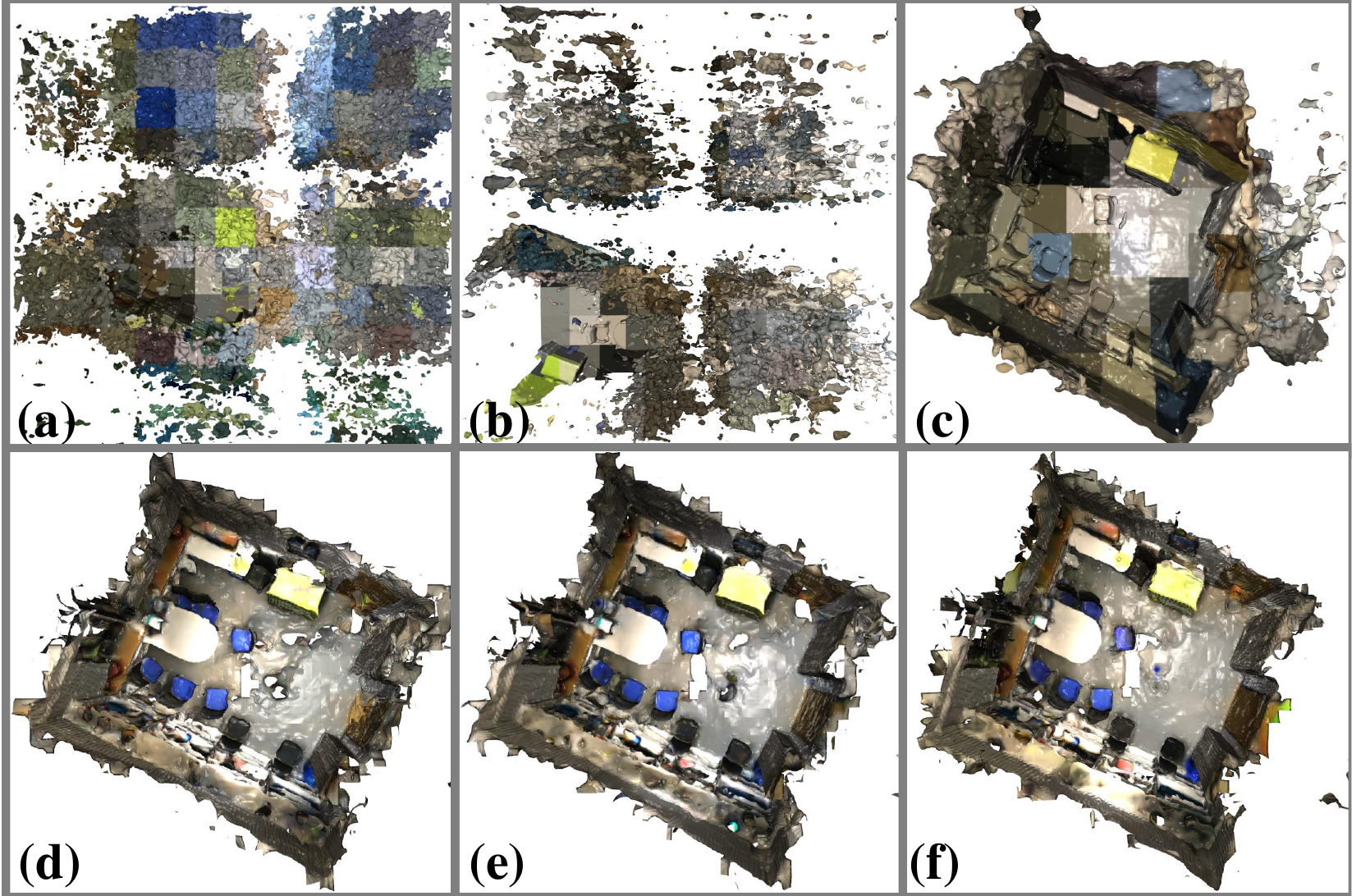}

    \caption{Reconstructed results of \cite{wang2023co} in scene0525 with inaccurate scene bounds (abc). Compared to degraded mesh quality and anamorphic patch-like texture, Ours (f) can perform high-fidelity mapping without any boundary priors; Compared to SVO-based baseline \cite{yang2022vox} (d) and LC-Fusion (e), Ours (f) demonstrate more consistent reconstructions.}
    \label{bound_comp}
    \vspace{-1.5em}
\end{figure}

\subsection{Evaluation of Scene Representation} \label{scene bounds}
1) \textit{Independence on Scene Bounds}: Since bounded neural implicit SLAM systems \cite{wang2023co} require scene bounds as input for joint encoding and mesh extraction \cite{lorensen1987marching}, we provide inaccurate scene bounds for \cite{wang2023co} to magnify their reliance on scene bounds and essential difference with our LCP-Fusion. The reconstructed results given different scene bounds are shown in the first row of Fig. \ref{bound_comp}: (a) oversized for encoding and marching cubes; (b) undersized for both; (c) inaccurate for encoding but accurate for marching cubes; (f) LCP-Fusion (Ours) without boundary prior. As can be seen, inaccurate scene bounds for both in \cite{wang2023co} will contribute to terrible mesh results. Moreover, even with accurate bounds for marching cubes, the surface texture still distorts due to the inappropriate encoding scale. For LCP-Fusion using SVO-based scene representation, we encode scenes and extract surfaces within dynamically allocated hybrid voxels, making it more suitable for scenes with unknown boundaries.

2) \textit{Introduction of Hybrid representation}: As shown in earlier results, the SVO-based baseline \cite{yang2022vox} tends to suffer localization drift and inconsistent mapping in real scenes with potential loop closure, which is attributed to the lack of local constraints and difficulties in high-dimensional correction. Compared to (d) Vox-Fusion* and (e) LC-Fusion (Ours w/o hybrid rep.), (f) LCP-Fusion (Ours) yields more consistent reconstructions, which is shown in the second row of Fig. \ref{bound_comp}. Moreover, Table. \ref{table 4} presents quantitative comparisons with LC-Fusion that demonstrate improved localization accuracy using our hybrid representation. Therefore, in addition to our proposed local constraints, the hybrid scene representation will bring further enhancements to tracking and mapping.

\begin{figure*}
    \centering
    \includegraphics[width=\linewidth]{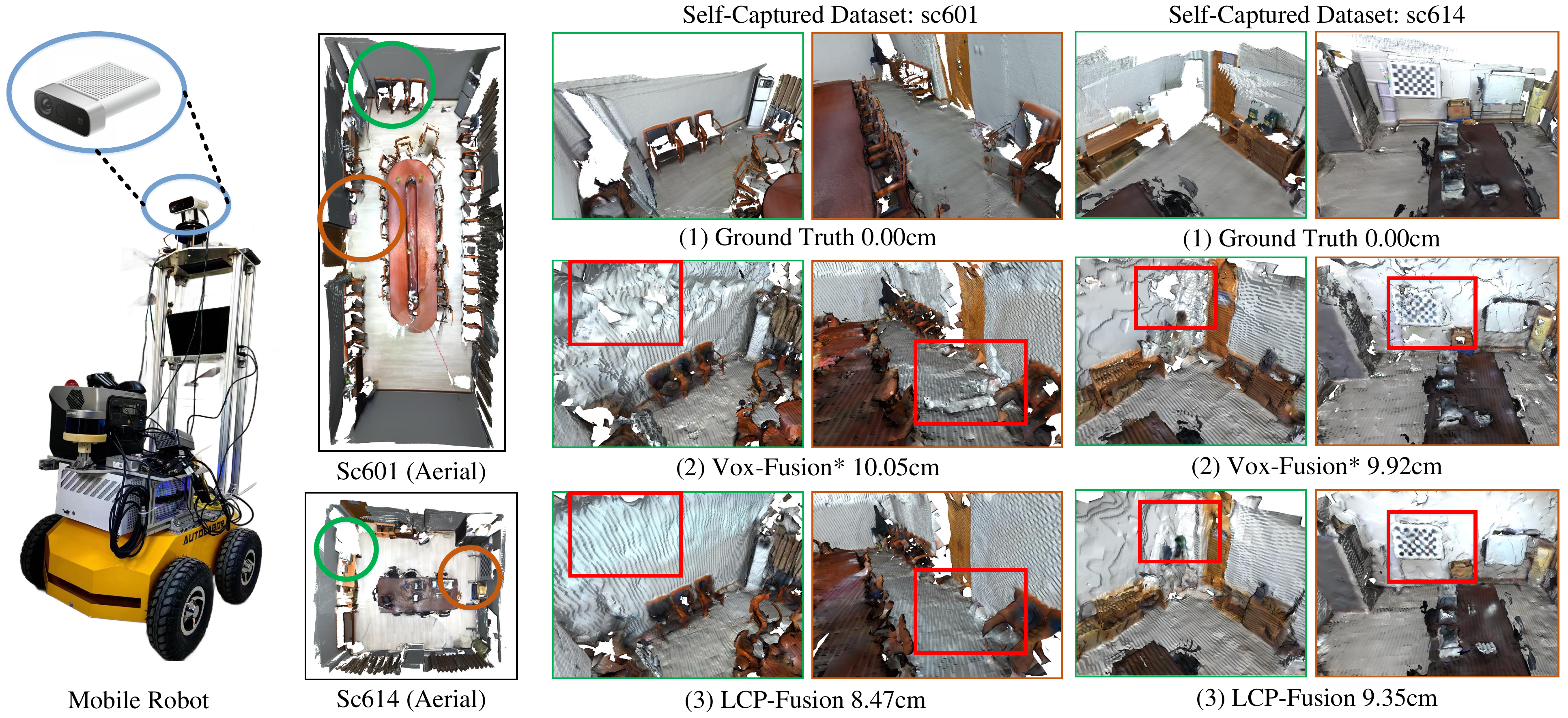}
    \vspace{-1.0em}
    \caption{Reconstructed mesh results of our self-captured dataset: sc601 and sc614. With improved localization accuracy, the highlight regions of zoomed windows indicate that our method yields more consistent surface against \cite{yang2022vox}. }
    \label{self_captured}
    \vspace{-1.0em}
\end{figure*}

\subsection{Ablation Study} \label{exp_ablation}
1) \textit{Effect of enhanced constraints and priors}: In order to evaluate the effectiveness of all designed components, we conduct an ablation study on a representative scene, the quantitative results are shown in Table. \ref{table 3}. As can be seen, our system with both local constraints and priors achieves highest localization accuracy among baseline and our incomplete variants:  (LCr) LCP-Fusion with only our window selection; (LCw) LCP-Fusion with only our warping loss; (LC) LCP-Fusion without SDF priors.
\begin{table}[t]
    \centering
    \caption{Ablation study on scene0181 with baseline and variants}
    \label{table 3}
    \begin{tabular}{cccccc}
    \toprule
         \multirow{2}{*}{Name}&\multicolumn{2}{c}{window selection}&\multirow{2}{0.1\linewidth}{\centering Warp\\loss}&\multirow{2}{0.1\linewidth}{\centering SDF\\prior}&\multirow{2}{*}{ATE (cm)}  \\
         \cmidrule(r){2-3}
         & random & ours \\
          \midrule
          Vox-Fusion*& \(\surd\)& & & &15.38\\
          LCr-Fusion& &\(\surd\) & & &13.99\\
          LCw-Fusion&\(\surd\)& &\(\surd\)& & 14.21\\
          LC-Fusion& &\(\surd\)&\(\surd\)& &11.19\\
          LCP-Fusion& &\(\surd\)&\(\surd\)&\(\surd\)&\textbf{10.18}\\
         \bottomrule
    \end{tabular}
    \vspace{-1.0em}
\end{table}
\begin{figure}
    \centering
    \includegraphics[width=\linewidth]{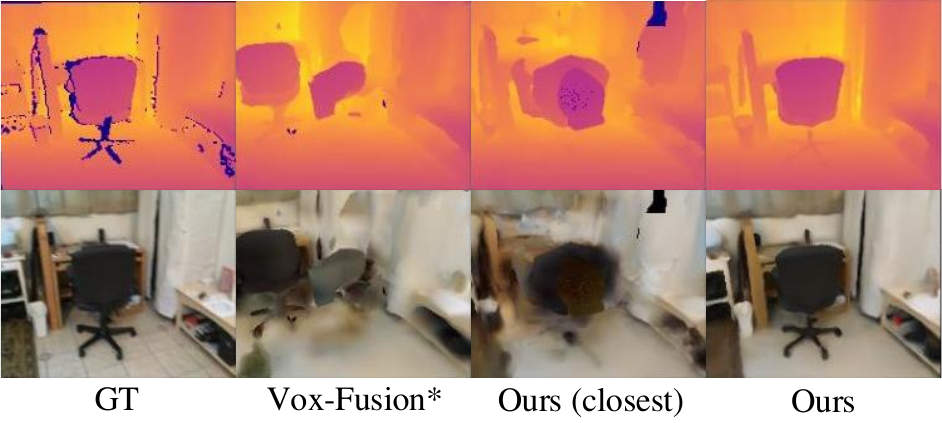}
    \caption{Rendered results in loop-closure regions of scene0000. Ours (closest) selects local keyframes in sliding window with most visual overlap; Ours selects randomly in a series of local keyframes with visual overlap.}
    \label{fig:ablation1}
    \vspace{-1.0em}
\end{figure}

\begin{figure}[ht]
\vspace{-0.25em}
    \centering
    \includegraphics[width=0.95\linewidth]{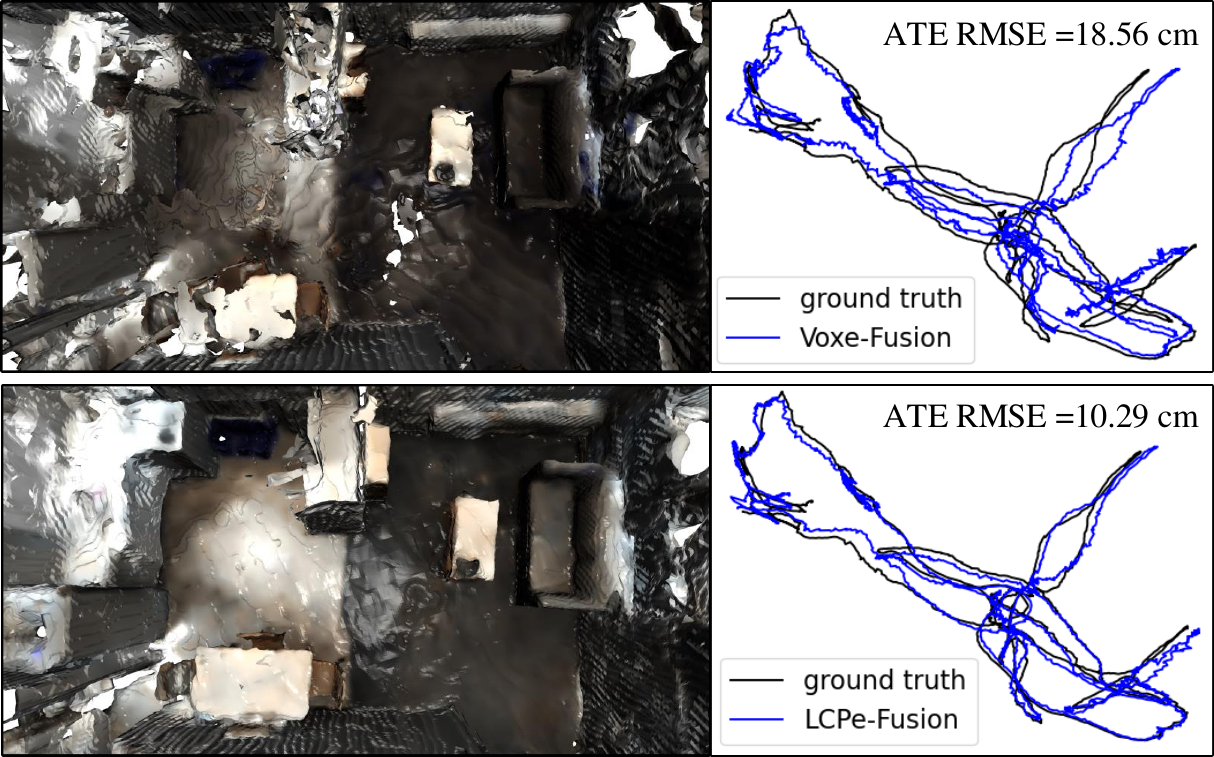}
  \vspace{-0.5em}
    \caption{SLAM results with adaptive early ending on scene0181. Compared with theoretical iterations of BA, Voxe-Fusion*(top) is trained for 72.99\% iterations, while LCPe-Fusion (bottom) for 69.94\%. Owing to the introduction of SDF priors, our system still perform stably with less iterations. }
    \label{fig:ablation2}
    \vspace{-1.0em}
\end{figure}

2) \textit{Effect of SDF priors with early ending}: Utilizing SDF priors for reasonable initialization, we introduce an early ending policy for accelerating mapping process without dramatic degradation in performance. Specifically, current iterative optimization breaks in advance if there are over \({Iter}_m/3\) losses less than average loss on previous frames. Quantitative results for time and accuracy are shown in Table. \ref{table 4} and qualitative comparison in Fig. \ref{fig:ablation2}. It is evident that, despite the adaptive early ending, our system (LCPe-Fusion) still outperforms the main baselines \cite{yang2022vox} in terms of localization accuracy and reconstruction consistency ,with a speed comparable to \cite{yang2022vox} and accuracy equivalent to our variants without SDF priors (LC-Fusion).
\begin{table}[t]
\setlength{\tabcolsep}{3.8pt}
\caption{ATE RMSE (cm)  and Time consumption (min) Ablations on selected ScanNet sequences}
\label{table 4}
  \centering
    \begin{tabular}{cccclcccc}
   \toprule
   Scene ID&  0059 & 0169 & 0181   &0207& 0472 & 0525& Avg.& Time\\
   \midrule
   Vox-Fusion*  & 9.06 & 9.64 & 15.38   &7.74& 9.98& 8.42& 10.037& \textbf{87.4} \\
   LC-Fusion &  \underline{8.13} & 6.39 & \underline{11.19}   &\underline{6.53}& 9.30& \underline{5.49}& \underline{7.838}& 96.2 \\
   LCPe-Fusion & 8.29 & \underline{5.99} & 11.71   &6.61& \underline{8.70}& 5.75& 7.842& \underline{87.8}\\
    LCP-Fusion&  \textbf{7.56}& \textbf{5.91}& \textbf{10.18} &\textbf{6.29}& \textbf{8.33}& \textbf{5.06}& \textbf{7.222}& 103.2\\
   \bottomrule
\end{tabular}
\vspace{-1.5em}
\end{table}

\section{CONCLUSIONS}

We propose LCP-Fusion (a neural implicit SLAM system with enhanced \textbf{L}ocal \textbf{C}onstraints and \textbf{C}omputable \textbf{P}rior). Utilizing the SVO-based hybrid scene representation, we show that jointly optimizing scene representation and poses in a novel sliding window consisting of local overlapped and historical keyframes, as well as constraining relative poses and geometry with warping loss, achieves accurate localization and consistent reconstruction in real scenes with noise and potential loop-closure. Furthermore, our introduction of computable SDF prior provides reasonable initialization for parametric encoding to further improve and stabilize performance even when mapping iteration decreases. Compared to our baselines, we've observed a 28.1\% improvement in localization accuracy on real datasets and 10.6\% on synthetic datasets, further validated under our self-captured dataset.  However, LCP-Fusion is still limited to basic spatial understanding of geometry and color from RGB-D inputs. Noting the impressive advances in web-pretrained visual language models (VLMs), neural implicit visual language SLAM for robotic downstream tasks can be our future work.




\bibliographystyle{Bibliography/IEEEtran}
\bibliography{Bibliography/IROS}

\end{document}